\documentclass[10pt,twocolumn,letterpaper]{article}

\usepackage{cvpr}              
\definecolor{cvprblue}{rgb}{0.21,0.49,0.74}
\usepackage[pagebackref,breaklinks,colorlinks,allcolors=cvprblue]{hyperref}
\usepackage{algorithm}
\usepackage{algorithmic}
\usepackage{tikz}
\usetikzlibrary{positioning,shapes.geometric,fit,arrows.meta,calc}
\usepackage[table]{xcolor}
\newcommand{\method}{PiJEPA}
\newcommand{\ours}{\method{}}
\newcommand{\R}{\mathbb{R}}

\newcommand{\cL}{\mathcal{L}}

\newcommand{\cN}{\mathcal{N}}
\newcommand{\cE}{\mathcal{E}}

\def\paperID{*****} 
\def\confName{CVPR}
\def\confYear{2026}

\title{Policy-Guided World Model Planning\\for Language-Conditioned Visual Navigation}

\author{Amirhosein Chahe ~~~~~~ Lifeng Zhou\thanks{Corresponding author.}\\
Drexel University\\
Philadelphia PA 19104, USA\\
{\tt\small \{ac4462, lz457\}@drexel.edu}
}

\begin{document}

\maketitle

\begin{abstract}
Navigating to a visually specified goal given natural language instructions remains a fundamental challenge in embodied AI.
Existing approaches either rely on reactive policies that struggle with long-horizon planning, or employ world models that suffer from poor action initialization in high-dimensional spaces.
We present \ours{}, a two-stage framework that combines the strengths of learned navigation policies with latent world model planning for instruction-conditioned visual navigation.
In the first stage, we finetune an Octo-based generalist policy, augmented with a frozen pretrained vision encoder (DINOv2 or V-JEPA-2), on the CAST navigation dataset to produce an informed action distribution conditioned on the current observation and language instruction.
In the second stage, we use this policy-derived distribution to warm-start Model Predictive Path Integral (MPPI) planning over a separately trained JEPA world model, which predicts future latent states in the embedding space of the same frozen encoder.
By initializing the MPPI sampling distribution from the policy prior rather than from an uninformed Gaussian, our planner converges faster to high-quality action sequences that reach the goal.
We systematically study the effect of the vision encoder backbone, comparing DINOv2 and V-JEPA-2, across both the policy and world model components.
Experiments on real-world navigation tasks demonstrate that \ours{} significantly outperforms both standalone policy execution and uninformed world model planning, achieving improved goal-reaching accuracy and instruction-following fidelity. Code is available at: \href{https://github.com/AmirhoseinCh/PiJEPA.git}{https://github.com/AmirhoseinCh/PiJEPA}.
\end{abstract}

\section{Introduction}
\label{sec:intro}

Building autonomous agents that can navigate to a goal specified by an image and a natural language instruction is a long-standing challenge in robotics and computer vision~\cite{zhu2017target,anderson2018vision,shah2023vint}.
Given a current egocentric observation $o_t$, a goal image $o_g$, and an instruction $\ell$ (\eg ``move towards the stairs''), the agent must produce a sequence of actions that reliably reaches the goal while respecting the semantics of the instruction.

Recent vision-language-action (VLA) models~\cite{octo2024,brohan2023rt2,shah2023vint,glossop2025cast} have made remarkable strides toward this objective.
Generalist policies such as Octo~\cite{octo2024} and NoMaD~\cite{sridhar2023nomad} can be finetuned on domain-specific navigation data to achieve impressive short-horizon control, while the CAST framework~\cite{glossop2025cast} further strengthens instruction following through counterfactual language-action augmentation.
Despite their effectiveness, these reactive policies predict actions in a single forward pass and thus lack the capacity to reason about the long-term consequences of their decisions.

A complementary line of work addresses this limitation by learning world models that predict future states conditioned on candidate actions, enabling explicit planning toward the goal~\cite{lecun2022path,hafner2024dreamerv3,hansen2024tdmpc2,bar2025nwm,zhou2024dinowm}.
Among these, Joint-Embedding Predictive Architecture World Models (JEPA-WMs)~\cite{terver2025jepawms,zhou2024dinowm,assran2025vjepa2ac} are especially attractive: by learning dynamics in the latent space of a frozen pretrained encoder, they support efficient rollouts without pixel-level reconstruction.
At inference time, a sampling-based optimizer such as MPPI~\cite{williams2015mppi,williams2017mppi} or CEM~\cite{rubinstein1999cem} searches over action sequences by unrolling the learned predictor and minimizing the embedding-space distance to the encoded goal.
However, this search must typically begin from an uninformed prior over a high-dimensional action space, resulting in slow convergence and susceptibility to local minima~\cite{terver2025jepawms}.

We propose \ours{}, a framework that bridges these two paradigms by using a learned policy to \emph{warm-start} world model planning.
Our key insight is that a finetuned VLA policy, while insufficient for long-horizon reasoning on its own, provides a highly informative \emph{action prior}: it concentrates probability mass in the region of action space most likely to make progress toward the goal.
Initializing the MPPI sampling distribution from this prior allows the world model planner to devote its search budget to refining already-promising trajectories rather than exploring the full action space from scratch.

Concretely, our pipeline operates as follows (see Figure~\ref{fig:overview}):
\begin{enumerate}
    \item \textbf{Policy prior.} Given the current observation $o_t$ and instruction $\ell$, a finetuned Octo model produces $N_\pi$ action chunk samples via its diffusion head. We transform these into the world model's local-frame action space and compute their empirical mean $\mu_\pi$ and standard deviation $\sigma_\pi$.
    \item \textbf{MPPI planning.} A JEPA world model, trained with the same frozen encoder, predicts future latent states. We run MPPI initialized at $(\mu_\pi, \sigma_\pi)$ and optimize against the embedding-space distance to the encoded goal $o_g$.
    \item \textbf{Execution.} The first action of the optimized sequence is executed, and the process repeats at the next replanning step.
\end{enumerate}
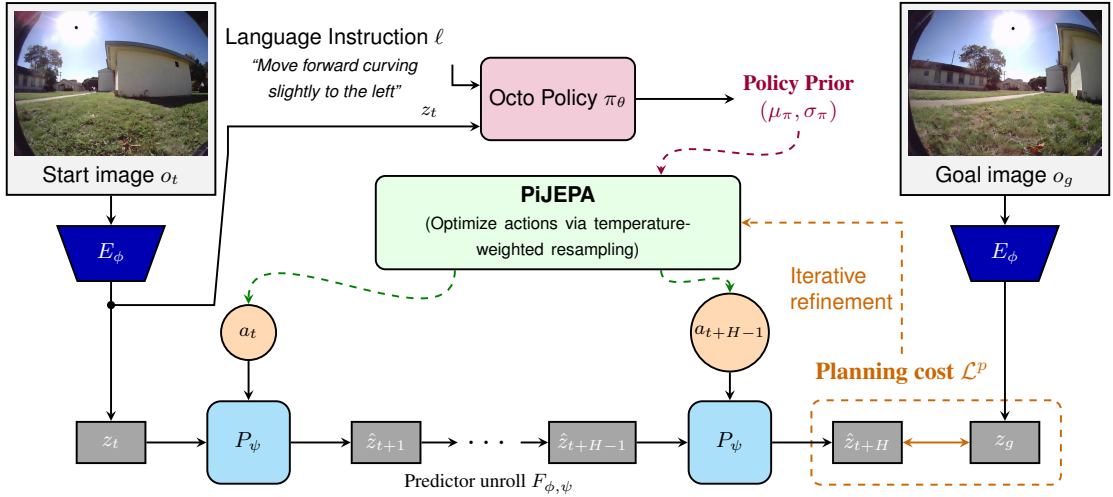
\begin{figure*}[ht]
    \centering
    \resizebox{0.85\textwidth}{!}{
    \begin{tikzpicture}[
        >=stealth,
        font=\small\sffamily,
        imgnode/.style={rectangle, draw, thick, minimum width=2.8cm, minimum height=1.8cm, text centered, fill=gray!10, align=center},
        enc/.style={trapezium, trapezium left angle=110, trapezium right angle=110, draw, thick, fill=blue!70!black, text=white, minimum height=0.8cm, text centered},
        latent/.style={rectangle, draw, thick, fill=gray!70, text=white, minimum width=1.0cm, minimum height=0.6cm, text centered},
        pred/.style={rectangle, draw, thick, rounded corners=0.2cm, fill=cyan!30, minimum width=1.2cm, minimum height=1.2cm, text centered},
        action/.style={circle, draw, thick, fill=orange!30, minimum size=0.8cm, inner sep=1pt, text centered},
        policy/.style={rectangle, draw, thick, rounded corners, fill=purple!20, minimum width=2.2cm, minimum height=1.2cm, text centered, align=center},
        mppi/.style={rectangle, draw, thick, rounded corners, fill=green!10, text width=5cm, minimum height=0.8cm, text centered, inner sep=0.15cm},
        txt/.style={text centered, align=center}
    ]

        \node[imgnode] (ot) at (0, 5.0) {\includegraphics[width=2.8cm]{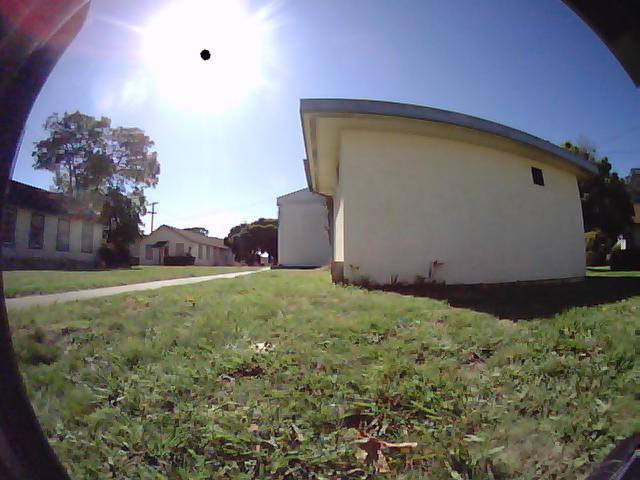} \\ Start image $o_t$};
        \node[txt] (inst) at (3.25, 5.50) {Language Instruction $\ell$\\\scriptsize{\textit{``Move forward curving}}\\\scriptsize{\textit{slightly to the left''}}};
        \node[policy] (octo) at (6.5, 5.0) {Octo Policy $\pi_\theta$};
        \node[txt, font=\bfseries\small, text=purple!80!black] (stats) at (10.0, 5.0) {Policy Prior\\$(\mu_\pi, \sigma_\pi)$};
        \node[imgnode] (og) at (13, 5.0) {\includegraphics[width=2.8cm]{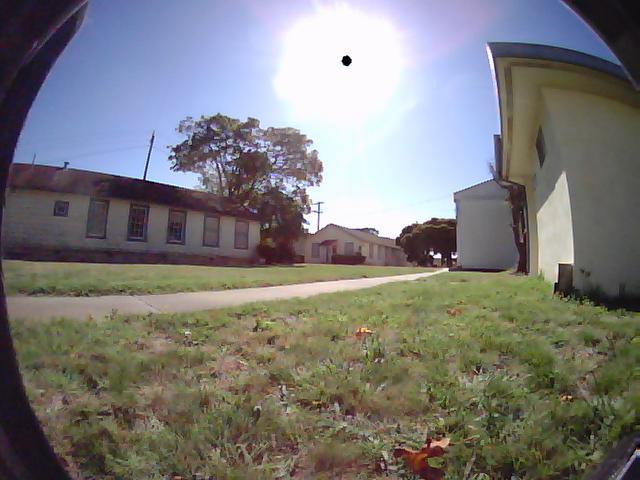} \\ Goal image $o_g$};

        \node[enc] (e1) at (0, 2.75) {$E_\phi$};
        \node[mppi] (mppi) at (6.5, 3.2) {\textbf{\ours}\\ \scriptsize(Optimize actions via temperature-weighted resampling)};
        \node[enc] (e2) at (13, 2.75) {$E_\phi$};

        \node[action, font=\footnotesize] (a1) at (2, 1.6) {$a_t$};
        \node[action, font=\footnotesize] (ah) at (9, 1.6) {$a_{t+H-1}$};

        \node[latent] (zt) at (0, 0) {$z_t$};
        \node[pred] (p1) at (2, 0) {$P_\psi$};
        \node[latent] (z1) at (4, 0) {$\hat{z}_{t+1}$};
        \node[txt] (dots) at (5.5, 0) {\Large $\dots$};
        \node[txt, font=\footnotesize] (unroll) at (5.5, -0.6) {Predictor unroll $F_{\phi, \psi}$};
        \node[latent] (zh1) at (7, 0) {$\hat{z}_{t+H-1}$};
        \node[pred] (ph) at (9, 0) {$P_\psi$};
        \node[latent] (zh) at (11, 0) {$\hat{z}_{t+H}$};
        \node[latent] (zg) at (13, 0) {$z_g$};

        \node[rectangle, draw=orange!80!black, dashed, thick, rounded corners, inner sep=0.3cm, fit=(zh) (zg)] (costbox) {};
        \node[above=0.1cm of costbox, xshift=-0.5cm, text=orange!80!black, font=\bfseries] (costlabel) {Planning cost $\cL^p$};
        \draw[<->, thick, orange!80!black] (zh) -- (zg);


        \draw[->, thick] (ot) -- (e1);
        \draw[->, thick] (e1) -- (zt);
        \draw[->, thick] (og) -- (e2);
        \draw[->, thick] (e2) -- (zg);

        \draw[->, thick] (0, 2.0) node[circle, fill, inner sep=1.2pt] {} 
            -- (1.5, 2) 
            -- (1.7, 4.2) 
            |- ([yshift=-0.4cm]octo.west) node[pos=0.9, above, font=\footnotesize] {$z_t$};

        \draw[->, thick] (inst.east) |- ([yshift=0.2cm]octo.west);
        \draw[->, thick] (octo.east) -- (stats.west);

        \draw[->, thick, dashed, purple!80!black] (stats.south) to[out=270, in=90] ([xshift=1.5cm]mppi.north) node[midway, right=0.1cm] {};
        
        \draw[->, thick, dashed, orange!80!black] (costlabel.north) |- (mppi.east) node[pos=0.25, left, align=left] {Iterative\\refinement};

        \draw[->, thick, dashed, green!50!black] ([xshift=-1.5cm]mppi.south) to[out=270, in=90] (a1.north);
        \draw[->, thick, dashed, green!50!black] ([xshift=1.5cm]mppi.south) to[out=270, in=90] (ah.north);

        \draw[->, thick] (a1.south) -- (p1.north);
        \draw[->, thick] (ah.south) -- (ph.north);

        \draw[->, thick] (zt) -- (p1);
        \draw[->, thick] (p1) -- (z1);
        \draw[->, thick] (z1) -- (dots);
        \draw[->, thick] (dots) -- (zh1);
        \draw[->, thick] (zh1) -- (ph);
        \draw[->, thick] (ph) -- (zh);

    \end{tikzpicture}
    }
    \caption{\textbf{Overview of \ours{}.} \textbf{Top:} The Octo policy, finetuned with a frozen vision encoder ($E_\phi$), takes the current latent observation $z_t$ and instruction $\ell$ as input and produces action chunk samples via its diffusion head. These are transformed from the global frame to the world model's local body frame. \textbf{Middle:} The policy's statistics $(\mu_\pi, \sigma_\pi)$ warm-starts MPPI, which iteratively optimizes the action distribution over $J$ iterations. \textbf{Bottom:} The JEPA world model predictor ($P_\psi$), trained with the same frozen encoder, autoregressively predicts future latent states. The MPPI candidates are scored by unrolling the world model and evaluating the latent-space distance to the encoded goal $z_g$ (Algorithm~\ref{alg:octo_mppi}).}
    \label{fig:overview}
\end{figure*}
Our contributions are:
\begin{enumerate}
    \item We propose \ours{}, a unified framework for language-conditioned visual navigation that combines a finetuned Octo policy with MPPI-based planning over a JEPA world model, using the policy output to warm-start the planner.
    \item We train both components with modern pretrained vision encoders (DINOv2 and V-JEPA-2), providing a systematic comparison of image-based vs.\ video-based representations for policy learning and world modeling in navigation.
    \item We demonstrate on real-world navigation tasks that policy-guided planning significantly outperforms both standalone policy execution and uninformed MPPI planning, improving goal-reaching accuracy and instruction-following fidelity.
\end{enumerate}

\section{Related Work}
\label{sec:related}

Our work draws on two active research threads, language-conditioned policies and latent world models; and unifies them through policy-guided planning. We review each in turn.

\subsection{Language-Conditioned Navigation Policies}

The integration of natural language with embodied perception has given rise to Vision-Language-Action (VLA) models that map instructions and visual observations directly to motor commands.
Generalist architectures such as Octo~\cite{octo2024} learn broadly from diverse robot data, while recent scalable designs including Dita~\cite{Hou2025DitaSD}, LAPA~\cite{Ye2024LatentAP}, and CLAP~\cite{Zhang2026CLAPCL} push the frontier of reactive instruction-following through latent action representations.
In the navigation domain specifically, UrbanNav~\cite{Mei2025UrbanNavLL} extends language-guided policies to complex urban environments via web-scale trajectory learning.
Although these systems achieve strong short-horizon performance, they share a common limitation: actions are selected greedily without explicit reasoning about future states, which degrades performance on tasks that require multi-step spatial planning.
\ours{} retains the rapid, instruction-grounded action proposals of a reactive VLA policy but uses them as a warm start for a separate planning stage, thereby combining language understanding with long-horizon deliberation.

\subsection{Latent World Models for Planning}

World models offer a principled route to long-horizon reasoning by enabling agents to simulate the consequences of candidate actions before committing to them.
Early approaches such as Navigation World Models (NWM)~\cite{bar2025nwm} synthesize pixel-level futures, but predicting dynamics in a learned latent space drastically reduces computational cost.
Joint-Embedding Predictive Architecture (JEPA) world models built on frozen foundation encoders like DINOv2 have proven particularly effective for physical planning tasks~\cite{zhou2024dinowm, terver2025jepawms}.
This family of models has been further advanced through continuous latent action formulations~\cite{Garrido2026LearningLA}, principled regularization strategies such as LeJEPA~\cite{Balestriero2025LeJEPAPA}, and joint VLA-latent pretraining~\cite{Sun2026VLAJEPAEV}.
Complementary efforts have also targeted the efficiency of the planning loop itself; for instance, One-Step World Model~\cite{Shen2026AnEA} reduces the computational latency of multi-step rollouts, and benchmarks like Target-Bench~\cite{Wang2025TargetBenchCW} evaluate world-model-based path planning toward text-specified targets.
Despite these advances, sampling-based planners such as MPPI and CEM remain sensitive to their initialization: starting from an uninformed Gaussian prior over high-dimensional action spaces leads to slow convergence and frequent entrapment in local minima~\cite{terver2025jepawms}.
\ours{} directly addresses this inference-time bottleneck.
Rather than conditioning the world model itself on language, which would conflate representation learning with instruction grounding, we keep the latent dynamics model language-agnostic and instead anchor the MPPI optimization with an instruction-aware action prior drawn from the VLA policy.
This decoupled design lets each component focus on what it does best: the policy provides efficient, semantically informed proposals, while the world model evaluates and refines them over extended horizons.
\section{Method}
\label{sec:method}

We present \ours{}, a two-stage approach for instruction-conditioned visual goal navigation.
An overview is shown in Figure~\ref{fig:overview}.
Our pipeline consists of three learned components: a frozen pretrained vision encoder, an Octo-based policy that generates an informed action prior, and a JEPA world model over which we perform MPPI planning warm-started by the policy prior.

\subsection{Problem Formulation}
\label{sec:formulation}

At each timestep $t$, the agent observes an egocentric RGB image $o_t$, is given a goal image $o_g$ and a natural language instruction $\ell$, and must produce a sequence of navigation actions $a_{1:H} \in \R^{H \times A}$ over a planning horizon $H$ that guides the robot towards the goal while respecting the instruction semantics.

\subsection{Policy Prior from Octo}
\label{sec:policy}

The first stage of \ours{} extracts an action prior from a finetuned Octo policy~\cite{octo2024}.
Octo is a transformer-based generalist policy with a diffusion action head that models the distribution $\pi_\theta(a | o_t, \ell)$ over action chunks.

Both the policy and the world model operate on top of a shared frozen pretrained vision encoder $E_\phi$ that maps observations to latent tokens $z_t = E_\phi(o_t) \in \R^{n \times d}$, where $n$ is the number of spatial tokens and $d$ is the embedding dimension.
The encoder weights $\phi$ remain frozen throughout, ensuring both components share a consistent representation space.

\subsection{JEPA World Model}
\label{sec:worldmodel}

The second component is a Joint-Embedding Predictive World Model (JEPA-WM)~\cite{terver2025jepawms,zhou2024dinowm} that predicts future latent states given the current state and actions.
The world model pairs the same frozen encoder $E_\phi$ with a learnable predictor $P_\psi$ and action encoder $A_\psi$.
Given a context of encoded observations and actions, the predictor forecasts the next latent state:
\begin{equation}
    \hat{z}_{t+1} = P_\psi\!\left(z_{t-w:t},\; A_\psi(a_{t-w:t})\right).
    \label{eq:predictor}
\end{equation}
The predictor uses a causal attention mask so it can predict from all context lengths up to a maximum window $w$.
Actions are injected into the predictor via a conditioning mechanism at every layer to prevent vanishing action signals through depth~\cite{terver2025jepawms}.

The predictor is trained to minimize the MSE between predicted and target embeddings.
Following~\cite{terver2025jepawms}, we use a multi-step rollout loss to improve long-horizon accuracy:
\begin{equation}
    \cL_\text{total} = \sum_{k=1}^{K_\text{roll}} \frac{1}{B}\sum_{b=1}^{B} \left\| F_{\phi,\psi}(z^b_t, a^b_{t:t+k-1}) - z^b_{t+k} \right\|^2_2,
    \label{eq:wm_loss}
\end{equation}
where $F_{\phi,\psi}$ denotes the autoregressive unrolling of the predictor (feeding each predicted $\hat{z}_{t+j}$ as context for the next step), trained with truncated backpropagation through time.

\subsection{Policy-Guided MPPI Planning}
\label{sec:planning}

At test time, we combine the Octo policy prior with the JEPA world model through MPPI-based planning~\cite{williams2015mppi,williams2017mppi}.
We first draw $N_\pi$ action chunk samples from the policy diffusion head.
Because the policy and world model may operate in different action coordinate frames, each sample is transformed by a mapping $T$ into the world model frame (details in Section~\ref{sec:setup}).
From the transformed samples $\tilde{a}^{(i)} = T(a^{(i)})$, we compute the initial MPPI mean and standard deviation:
\begin{equation}
    \mu^0 = \frac{1}{N_\pi}\sum_{i=1}^{N_\pi} \tilde{a}^{(i)}, \quad
    \sigma^0 = \text{clamp}\!\left(\text{std}(\{\tilde{a}^{(i)}\}),\; \sigma_\text{min},\; \sigma_\text{max}\right),
    \label{eq:prior}
\end{equation}
which defines a policy-informed initialization over action sequences.

\noindent\textbf{Planning objective.}
Given the encoded current observation $z_t$ and goal encoding $z_g$, we evaluate a candidate action sequence $a_{1:H}$ by the terminal latent distance
\begin{equation}
    \cL^p(z_t, a_{1:H}, z_g) = \frac{1}{n}\sum_{i=1}^{n}\left\| \hat{z}^{(i)}_{t+H} - z^{(i)}_g \right\|^2_2,
    \label{eq:plan_cost}
\end{equation}
where $\hat{z}_{t+H} = F_{\phi,\psi}(z_t, a_{1:H})$ is obtained by autoregressively rolling out the world model for $H$ steps.

\noindent\textbf{MPPI refinement.}
Starting from $(\mu^0,\sigma^0)$, MPPI iteratively samples candidate action sequences, evaluates them using Eq.~\ref{eq:plan_cost}, and updates the sampling distribution using temperature-weighted elite trajectories.
At iteration $j$, elite weights are computed as
\begin{equation}
    w_k = \frac{\exp\!\big(\lambda\,(c_\text{min} - c_k)\big)}{\sum_{k'=1}^{K} \exp\!\big(\lambda\,(c_\text{min} - c_{k'})\big)},
    \label{eq:mppi_weights}
\end{equation}
and the Gaussian parameters are updated by
\begin{equation}
    \mu^{j+1} = \sum_{k} w_k\, a^{(k)}, \qquad
    (\sigma^{j+1})^2 = \sum_{k} w_k \big(a^{(k)} - \mu^{j+1}\big)^2,
    \label{eq:mppi_update}
\end{equation}
with $\sigma^{j+1}$ clamped to $[\sigma_\text{min}, \sigma_\text{max}]$.
After $J$ iterations, one elite trajectory is sampled proportionally to $w_k$ and executed.
The full procedure is summarized in Algorithm~\ref{alg:octo_mppi}.

\noindent\textbf{Conditioning.}
The language instruction $\ell$ enters only through the Octo policy, which shapes the initial planning distribution.
The world model is language-agnostic and models only visual-action dynamics.
The goal image $o_g$ is encoded into $z_g$ and enters through the planning cost in Eq.~\ref{eq:plan_cost}.

\begin{algorithm}[t]
\caption{\ours: Policy-Guided World Model MPPI Planning}
\label{alg:octo_mppi}
\begin{algorithmic}[1]
\REQUIRE Obs.\ $o_t$, goal $o_g$, instruction $\ell$, policy $\pi_\theta$, world model $P_\psi$, encoder $E_\phi$
\REQUIRE Horizon $H$, samples $N$, elites $K$, iterations $J$, temperature $\lambda$, policy samples $N_\pi$
\vspace{1.5mm}
\item[] \textbf{\textit{Stage 1: Policy prior}}
\STATE $z_t \leftarrow E_\phi(o_t)$, \quad $z_g \leftarrow E_\phi(o_g)$
\FOR{$i = 1, \ldots, N_\pi$}
    \STATE $a^{(i)} \sim \pi_\theta(\cdot \mid o_t, \ell)$ \hfill $\triangleright$ diffusion sampling
    \STATE $\tilde{a}^{(i)} \leftarrow T(a^{(i)})$ \hfill $\triangleright$ coordinate transform
\ENDFOR
\STATE $\mu^0 \leftarrow \text{mean}(\{\tilde{a}^{(i)}\})$ 
\STATE $\sigma^{0} \leftarrow  \text{clamp}(\text{std}(\{\tilde{a}^{(i)}\}), \sigma_\text{min}, \sigma_\text{max})$
\vspace{1.5mm}
\item[] \textbf{\textit{Stage 2: MPPI planning}}
\FOR{$j = 0, \ldots, J - 1$}
    \FOR{$i = 1, \ldots, N$}
        \STATE $a^{(i)}_{1:H} \leftarrow \text{clamp}\big(\mu^j + \sigma^j \odot \epsilon^{(i)},\; {-1},\; 1\big)$
        \STATE $\epsilon^{(i)}\!\sim\!\cN(0,I)$
        \STATE $\hat{z} \leftarrow z_t$
        \FOR{$h = 1, \ldots, H$}
            \STATE $\hat{z} \leftarrow P_\psi(\hat{z},\; a^{(i)}_h)$ \hfill $\triangleright$ autoregressive rollout
        \ENDFOR
        \STATE $c^{(i)} \leftarrow \| \hat{z} - z_g \|_2^2 / n$ \hfill $\triangleright$ Eq.~\ref{eq:plan_cost}
    \ENDFOR
    \STATE $\cE \leftarrow \text{top-}K\text{ by lowest } c^{(i)}$
    \STATE $w_k \leftarrow \exp(\lambda\,(c_\text{min} - c_k)) \;/\; \sum_{k'} \exp(\lambda\,(c_\text{min} - c_{k'}))$
    \STATE $\mu^{j+1} \leftarrow \sum_{k} w_k\, a^{(k)}$
    \STATE $\sigma^{j+1} \leftarrow \text{clamp}\!\left(\sqrt{\sum_k w_k (a^{(k)} \!-\! \mu^{j+1})^2},\; \sigma_\text{min},\; \sigma_\text{max}\right)$
\ENDFOR
\STATE Sample $k^* \in \cE$ with probability $w_{k^*}$
\RETURN $a^{(k^*)}_{1:H}$
\end{algorithmic}
\end{algorithm}

\section{Experiments}
\label{sec:experiments}

\subsection{Experimental Setup}
\label{sec:setup}

\noindent\textbf{Dataset.}
We train and evaluate on the CAST dataset~\cite{glossop2025cast}, a large-scale visual navigation dataset augmented with counterfactual instruction-action pairs.
CAST addresses the problem of posterior collapse---where the policy ignores the language instruction because the observation alone suffices to predict the action---by using a VLM to generate alternative instructions and an atomic policy to produce corresponding counterfactual action labels.
Each action is a 4-dimensional vector $a = (\Delta x, \Delta y, \sin\Delta\phi, \cos\Delta\phi)$ specifying the robot's displacement and heading change; the $(\sin\Delta\phi, \cos\Delta\phi)$ encoding avoids angle discontinuities at $\pm\pi$.
Actions are normalized to $[-1, 1]$ using bounds normalization based on the 1st and 99th percentile statistics.

\noindent\textbf{Vision encoders.}
We study two frozen pretrained encoder families:
\emph{DINOv2 ViT-S}~\cite{oquab2024dinov2}, a self-supervised image encoder with strong spatial and object segmentation features; and
\emph{V-JEPA-2 ViT-L}~\cite{assran2025vjepa2}, a self-supervised video encoder that captures temporal dynamics.
For V-JEPA-2, we follow the frame-duplication strategy of~\cite{terver2025jepawms}, encoding each frame as a duplicated 2-frame video.
We apply layer normalization to the encoder output to stabilize prediction targets and the planning cost landscape.

\noindent\textbf{Policy training.}
We finetune the Octo-Small model~\cite{octo2024} on CAST, replacing its original visual encoder with the frozen pretrained encoder $E_\phi$ followed by a learnable projection $W_\text{proj} \in \R^{d \times d_\text{octo}}$.
Language instructions are encoded with a pretrained T5 model~\cite{raffel2020t5} (16 tokens).
The diffusion action head is a 3-layer MLP with residual connections, trained with the DDPM objective~\cite{ho2020ddpm} using a cosine noise schedule.
Training uses the CAST-augmented data including both original and counterfactual trajectory segments.

\begin{figure*}[h]
    \centering
    \small

    \begin{minipage}[c]{0.35\textwidth}
        \centering
        \begin{tabular}{@{}c@{\hspace{4pt}}c@{}}
            \includegraphics[width=0.48\textwidth]{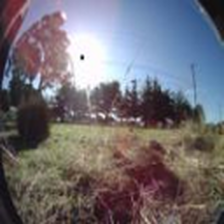} &
            \includegraphics[width=0.48\textwidth]{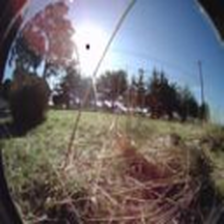} \\[-1pt]
            \footnotesize Start & \footnotesize Goal
        \end{tabular}

        \vspace{4pt}
        \textbf{Language instruction:}\textit{``Move forward with minor course correction''}
    \end{minipage}
    \begin{minipage}[c]{0.5\textwidth}
        \begin{subfigure}[t]{\textwidth}
            \centering
            \includegraphics[width=\textwidth]{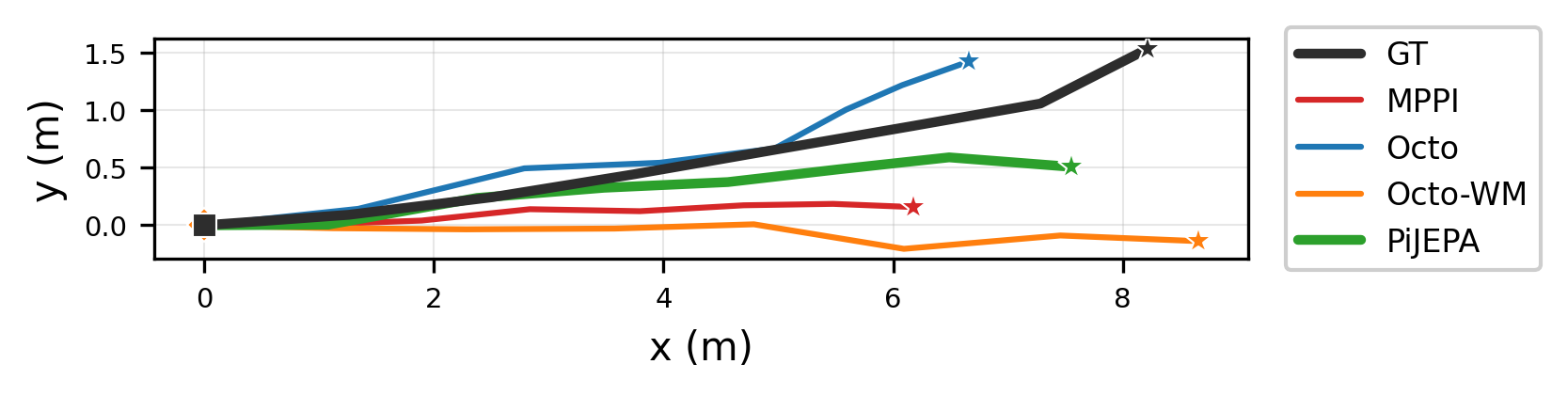}
            \caption{DINOv2 encoder}
            \label{fig:traj_dino}
        \end{subfigure}

        \vspace{4pt}

        \begin{subfigure}[t]{\textwidth}
            \centering
            \includegraphics[width=\textwidth]{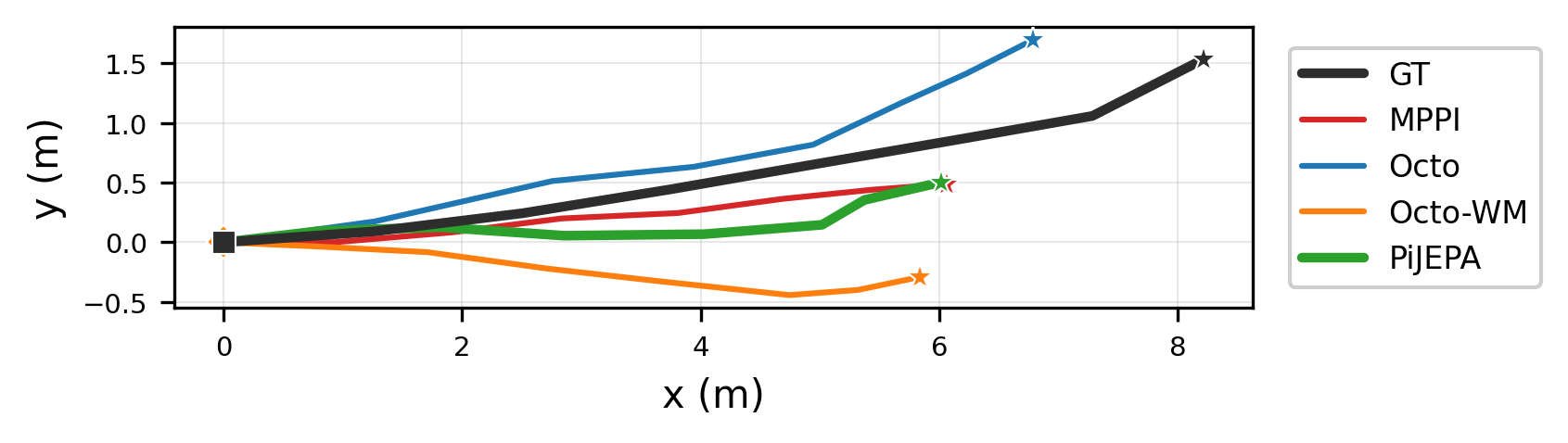}
            \caption{V-JEPA-2 encoder}
            \label{fig:traj_vjepa}
        \end{subfigure}
    \end{minipage}

    \caption{\textbf{Qualitative trajectory comparison.}
    Given a start observation, goal image, and language instruction (left), we compare trajectories produced by each method under two encoder backbones (right).
    The black curve (GT) shows the ground-truth path; colored curves show MPPI (red), Octo policy (blue), Octo-WM scoring (orange), and \method{} (green). Stars mark final positions.
    \method{} most closely tracks the ground truth in both settings.}
    \label{fig:qualitative_trajectory}
\end{figure*}

\noindent\textbf{Coordinate transform.}
The Octo policy outputs global-frame displacements, while the world model expects local body-frame actions.
We bridge this gap by accumulating headings from the $(\sin\Delta\phi, \cos\Delta\phi)$ components and rotating the displacement into the robot's local frame at each timestep:
\begin{equation}
    \begin{gathered}
        \phi_t = \textstyle\sum_{\tau=0}^{t-1} \text{atan2}(\sin\Delta\phi_\tau, \cos\Delta\phi_\tau), \\[2pt]
        \begin{pmatrix} \Delta x^\ell_t \\ \Delta y^\ell_t \end{pmatrix} = \begin{pmatrix} \cos\phi_t & \sin\phi_t \\ -\sin\phi_t & \cos\phi_t \end{pmatrix} \begin{pmatrix} \Delta x_t \\ \Delta y_t \end{pmatrix}.
    \end{gathered}
    \label{eq:coord_transform}
\end{equation}
The rotated actions are then normalized to $[-1, 1]$ using the dataset bounds statistics.

\begin{table*}[h!]
\centering
\caption{\textbf{DINOv2 ViT-S} on CAST validation episodes. Best in \textbf{bold}. All~$\downarrow$.}
\label{tab:dino}
\setlength{\tabcolsep}{2.5pt}
\small
\begin{tabular}{@{}l ccc ccc cc cc@{}}
\toprule
& \multicolumn{3}{c}{ATE XY\,(m)} & \multicolumn{3}{c}{ATE Hdg\,($^\circ$)} & \multicolumn{2}{c}{RPE XY\,(m)} & \multicolumn{2}{c@{}}{RPE Hdg\,($^\circ$)} \\
\cmidrule(lr){2-4}\cmidrule(lr){5-7}\cmidrule(lr){8-9}\cmidrule(l){10-11}
 & RMSE & Mean & Final & RMSE & Mean & Final & RMSE & Mean & RMSE & Mean \\
\midrule
Octo         & 1.98 & 1.65 & 3.19 & 20.08 & 16.76 & 28.55 & 0.61 & 0.57 &  8.29 &  6.94  \\
MPPI      & 1.85 & 1.48 & 3.23 & \textbf{18.79} & \textbf{15.68} & \textbf{27.86} & 0.55 & 0.51 & \textbf{5.70} & \textbf{5.11} \\
Octo-WM           & 1.80 & 1.43 & 3.15 & 22.16 & 18.60 & 29.16 & 0.56 & 0.51 & 7.51 & 6.45 \\
\rowcolor[gray]{.92}
\ours & \textbf{1.78} & \textbf{1.42} & \textbf{3.12} & 19.89 & 16.52 & 28.63 & \textbf{0.56} & \textbf{0.51} & 7.05 &  6.07 \\
\bottomrule
\end{tabular}
\end{table*}

\noindent\textbf{World model training.}
The JEPA world model predictor is a ViT with frame-causal attention and Adaptive Layer Normalization (AdaLN) action conditioning~\cite{peebles2023dit,terver2025jepawms}, which modulates scale and shift at every transformer block.
It is trained on CAST trajectory segments with the multi-step rollout loss (Eq.~\ref{eq:wm_loss}) using truncated backpropagation through time.
The encoder weights remain frozen; only the predictor and action encoder parameters are updated.
 All experiments were conducted on four NVIDIA H200 GPUs.

\noindent\textbf{Encoder variants.}
We train four model configurations to systematically study encoder choice:
(i)~DINOv2-Policy + DINOv2-WM, the most natural consistent-space pairing;
(ii)~V-JEPA-2-Policy + V-JEPA-2-WM, leveraging temporal features in both components;
(iii)~DINOv2-Policy + V-JEPA-2-WM; and
(iv)~V-JEPA-2-Policy + DINOv2-WM.
The cross-encoder configurations test whether the action prior from one representation space transfers to planning in another.
Prior work~\cite{terver2025jepawms} found DINOv2 excels at fine-grained spatial reasoning while V-JEPA-2 captures richer temporal structure; evaluating all four combinations reveals which properties matter for the policy prior versus latent dynamics.

\noindent\textbf{MPPI hyperparameters.}
We use $J=4$ MPPI iterations, $N=32$ candidate samples, $K=4$ elites, inverse temperature $\lambda=0.8$, and $N_\pi=4$ Octo diffusion samples for the policy prior.
The standard deviation is clamped to $[\sigma_\text{min}, \sigma_\text{max}] = [0.01, 0.05]$.

\subsection{Baselines and Evaluation}
\label{sec:baselines}

To isolate the contribution of each component, we compare four planning strategies:
\begin{itemize}
    \item \textbf{Default MPPI}: Standard MPPI initialized with $\mu^0 = \mathbf{0}$, $\sigma^0 = \sigma_\text{max}$, serving as a pure world model planning baseline.
    \item \textbf{\ours} (ours): Warm-started MPPI with $\mu^0 = \mu_\pi$, $\sigma^0 = \sigma_\pi$ from the policy prior (Algorithm~\ref{alg:octo_mppi}).
    \item \textbf{Octo-WM}: Draws $N_\pi$ samples from the policy, evaluates each via a full world model rollout, and selects the sample with the lowest cost $\cL^p$---using the world model for scoring without trajectory optimization.
    \item \textbf{Octo}: Octo predicted actions with no world model involvement, serving as a pure reactive policy baseline.
\end{itemize}

We report Absolute Trajectory Error (ATE) and Relative Pose Error (RPE) for both XY position (in meters) and heading (in degrees), computed between predicted and ground-truth trajectories on the CAST validation set.


\subsection{Results}
\label{sec:results}

Tables~\ref{tab:dino} and~\ref{tab:vjepa} report the full set of trajectory metrics for the DINOv2 and V-JEPA-2 encoder configurations, respectively.
We discuss the main findings below.

\begin{table*}[ht]
\centering
\caption{\textbf{V-JEPA-2 ViT-L} on CAST validation episodes. Best in \textbf{bold}. All~$\downarrow$.}
\label{tab:vjepa}
\setlength{\tabcolsep}{2.5pt}
\small
\begin{tabular}{@{}l ccc ccc cc cc@{}}
\toprule
& \multicolumn{3}{c}{ATE XY\,(m)} & \multicolumn{3}{c}{ATE Hdg\,($^\circ$)} & \multicolumn{2}{c}{RPE XY\,(m)} & \multicolumn{2}{c@{}}{RPE Hdg\,($^\circ$)} \\
\cmidrule(lr){2-4}\cmidrule(lr){5-7}\cmidrule(lr){8-9}\cmidrule(l){10-11}
 & RMSE & Mean & Final & RMSE & Mean & Final & RMSE & Mean & RMSE & Mean \\
\midrule
Octo         & 1.72 & 1.36 & 3.02 & 18.72 & 15.64 & 27.72 & 0.53 & 0.48 &  6.09 &  5.34  \\
MPPI    & 1.87 & 1.50 & 3.29 & 19.15 & 15.98 & 28.50 & 0.55 & 0.52 & \textbf{5.75} & \textbf{5.15} \\
Octo-WM           & 1.67 & 1.35 & \textbf{2.87} & 20.63 & 17.27 & 28.06 & 0.54 & 0.49 & 6.93 & 5.99 \\
\rowcolor[gray]{.92}
\ours    & \textbf{1.65} & \textbf{1.32} & 2.88 & \textbf{19.13} & \textbf{15.95} & \textbf{27.93} & \textbf{0.53} & \textbf{0.49} & 6.46 & 5.62 \\
\bottomrule
\end{tabular}
\end{table*}

\paragraph{\ours{} achieves the best positional accuracy across both encoders.}
\ours{} leads on ATE XY metrics in both configurations: 1.78\,m RMSE and 3.12\,m Final with DINOv2, and 1.65\,m RMSE and 1.32\,m Mean with V-JEPA-2, which is the lowest across all settings.
Octo-WM narrowly edges \ours{} on V-JEPA-2 Final position (2.87\,m vs.\ 2.88\,m), but \ours{} dominates on the remaining trajectory-level metrics, confirming that warm-starting MPPI with a policy prior enables the planner to refine already-promising trajectories rather than searching from scratch.

\paragraph{Uninformed MPPI struggles with position but excels at heading.}
MPPI exhibits a striking divergence across metric types: it is the weakest method on ATE XY under V-JEPA-2 (RMSE 1.87\,m, Final 3.29\,m), confirming that uninformed initialization wastes the sampling budget, yet it achieves the best heading metrics across both encoders (e.g., DINOv2: ATE Hdg RMSE 18.79$^\circ$, RPE Hdg Mean 5.11$^\circ$).
We attribute this to the embedding-space objective implicitly regularizing heading alignment even when translational accuracy is poor.

\paragraph{World model scoring captures much of the planning benefit.}
Octo-WM provides large positional gains over the raw Octo policy---with V-JEPA-2, ATE XY Final drops from 3.02\,m to 2.87\,m---suggesting that even without iterative optimization, using the world model to filter poor policy proposals is a powerful strategy.
The relatively small gap between Octo-WM and \ours{} indicates that filtering accounts for much of the benefit, though the additional MPPI refinement still yields the best overall trajectory accuracy.

\paragraph{The reactive policy provides strong local control.}
Despite lacking look-ahead, Octo remains competitive on step-level metrics: with V-JEPA-2 it achieves the best RPE XY Mean (0.48\,m) and strongest ATE Heading (RMSE 18.72$^\circ$, Final 27.72$^\circ$) as shown in Table~\ref{tab:vjepa}.
Planning thus contributes its main benefit at the trajectory level, where cumulative errors compound.
 PiJEPA and Octo are nearly indistinguishable on step-level RPE XY
(0.53 vs.\ 0.53\,m RMSE), yet PiJEPA opens a clear gap on trajectory-level ATE XY
(1.65 vs.\ 1.72\,m RMSE, 2.88 vs.\ 3.02\,m Final),
confirming that the planning stage prevents per-step errors from accumulating over the full horizon.

\paragraph{V-JEPA-2 yields stronger overall performance.}
V-JEPA-2 with \ours{} achieves the best overall positional accuracy, and even its Octo baseline (ATE XY RMSE 1.72\,m) surpasses the best DINOv2 method (1.78\,m).
However, uninformed MPPI performs worst under V-JEPA-2 (Final 3.29\,m vs.\ 3.23\,m for DINOv2), suggesting its richer latent dynamics create a harder optimization landscape for uninformed search.
Once anchored by the policy prior, this space becomes an asset: the gap between \ours{} and uninformed MPPI on ATE XY Final grows from 0.11\,m (DINOv2) to 0.41\,m (V-JEPA-2).

\paragraph{Planning latency.}
The Octo policy requires approximately 2.13\,s to generate its action proposals via diffusion sampling, while the MPPI planning stage adds only 0.35\,s, bringing the total \ours{} inference time to roughly 2.48\,s for an 8-step trajectory.
The lightweight MPPI overhead demonstrates that the planning stage is practical to layer on top of the policy, with the vast majority of latency attributable to the diffusion-based action sampling rather than the world model rollouts.

\begin{figure*}[ht!]
    \centering
    \small

    \textbf{Language instruction:} \textit{``Follow the building''}
    \vspace{6pt}

    \begin{minipage}[c]{0.62\textwidth}
        \centering
        \setlength{\tabcolsep}{1pt}
        \renewcommand{\arraystretch}{0.6}
        \begin{tabular}{@{}c c c c c c@{}}
            & \footnotesize Frame 0 & \footnotesize Frame 1 & \footnotesize Frame 3 & \footnotesize Frame 5 & \footnotesize Frame 7 \\[1pt]

            \rotatebox{90}{\footnotesize ~~~~~GT} &
            \includegraphics[width=0.19\linewidth]{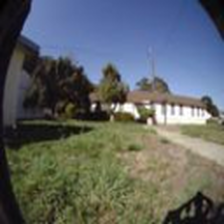} &
            \includegraphics[width=0.19\linewidth]{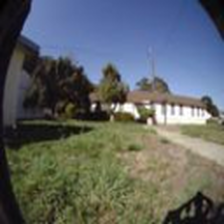} &
            \includegraphics[width=0.19\linewidth]{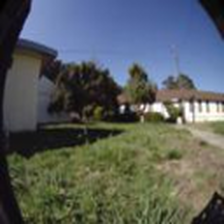} &
            \includegraphics[width=0.19\linewidth]{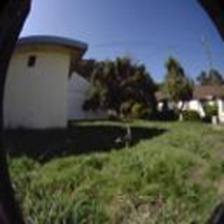} &
            \includegraphics[width=0.19\linewidth]{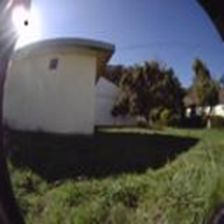} \\[2pt]

            \rotatebox{90}{\footnotesize ~~WM Pred.} &
            & 
            \includegraphics[width=0.19\linewidth]{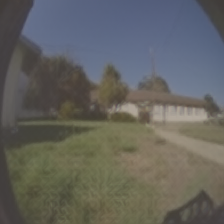} & 
            \includegraphics[width=0.19\linewidth]{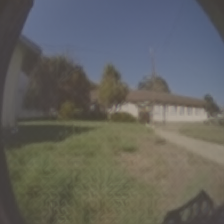} & 
            \includegraphics[width=0.19\linewidth]{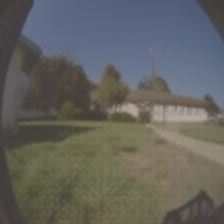} & 
            \includegraphics[width=0.19\linewidth]{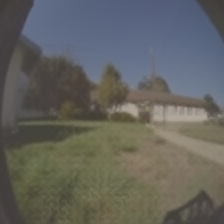} 
        \end{tabular}
    \end{minipage}
    \begin{minipage}[c]{0.37\textwidth}
        \centering
        \includegraphics[width=\linewidth]{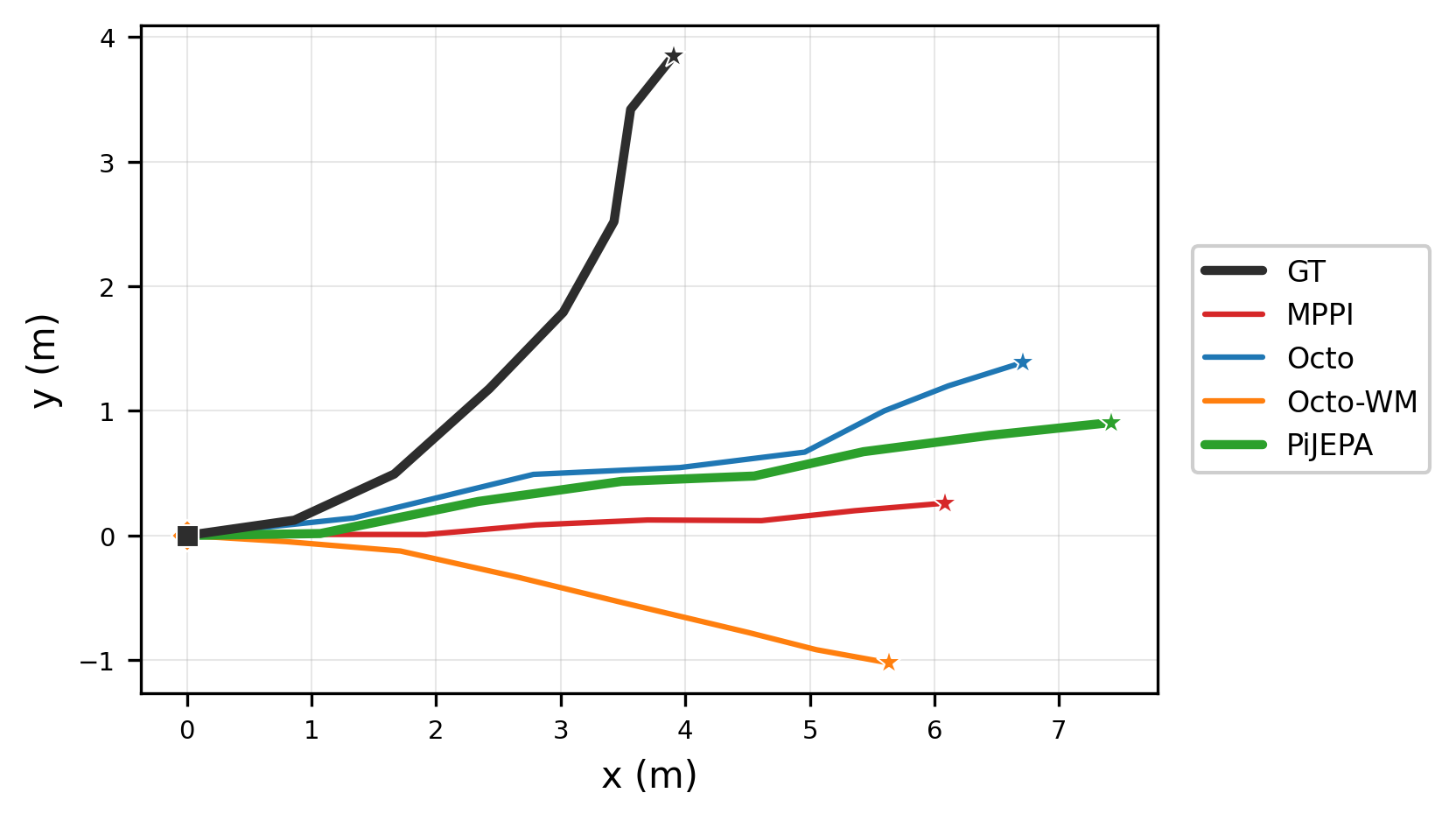}
    \end{minipage}

    \vspace{8pt} 
    \caption{\textbf{Failure case analysis.}
The language instruction \textit{``Follow the building''} is inherently ambiguous, as multiple buildings are visible in the scene.
The Octo policy (blue) misinterprets the referent and veers toward a different building, illustrating how vague instructions can mislead reactive policies that lack long-horizon reasoning.
Meanwhile, the world model fails to make meaningful progress because it becomes stuck in its rollouts, predicting nearly identical latent states, which causes the planner to stagnate.
The WM Pred.\ row confirms this directly. The predicted observations remain largely unchanged across the planning horizon.
\method{} (green) partially mitigates both issues by grounding the planner with a policy-derived prior, though it still undershoots the ground-truth trajectory.}
\label{fig:failure_case}
\end{figure*}

\paragraph{Qualitative results and failure analysis.}
Figure~\ref{fig:qualitative_trajectory} shows that \ours{} (green) most closely tracks the ground-truth path under both encoders, while MPPI (red) deviates substantially and Octo (blue) drifts over longer horizons.
Figure~\ref{fig:failure_case} presents a failure case with the ambiguous instruction \textit{``Follow the building,''} where the Octo policy veers toward the wrong building. The world model itself gets stuck in its rollouts, and the predicted latent states remain nearly identical across the planning horizon regardless of the sampled actions, leaving the planner unable to make progress.
\ours{} partially mitigates this through the policy prior, which bypasses the stalled rollouts by anchoring the search in a productive region of action space, though it still undershoots the ground truth, indicating that ambiguous instructions and world model stagnation remain challenging.

\section{Conclusion}
\label{sec:conclusion}

We have presented \ours{}, a framework for instruction-conditioned visual navigation that warm-starts MPPI planning over a JEPA world model using a finetuned VLA policy prior.
\ours{} achieves the best positional accuracy across both DINOv2 and V-JEPA-2 encoders, outperforming reactive policies, uninformed planning, and world model scoring baselines.
Our analysis reveals a natural division of labor: the reactive policy excels at local control and heading prediction, uninformed MPPI achieves strong heading alignment through its embedding-space objective, and world model scoring captures much of the look-ahead benefit by filtering poor proposals. \ours{} combines these strengths with the additional MPPI refinement, yielding the best trajectory-level coherence.
The MPPI planning latency overhead is negligible relative to the accuracy gains.
However, our failure analysis reveals that the world model can become stuck in its rollouts, producing static latent predictions that render the planner ineffective.
Addressing this stagnation through improved dynamics architectures or diversity-promoting rollout mechanisms is a key direction for future work, alongside intermediate waypoint costs, richer action spaces, and closed-loop replanning.

\section{Acknowledgment}

Computational resources were provided in part through NSF MRI Award Number 2320600.


{\small
\bibliographystyle{ieeetr}
\bibliography{references}
}

\end{document}